\documentclass[letterpaper, 10 pt, conference]{ieeeconf}  

\IEEEoverridecommandlockouts                              
\usepackage{graphics} 
\usepackage{graphicx}
\usepackage{xcolor}
\usepackage{amsmath}
\usepackage{mathtools}
\usepackage{cite}
\usepackage{caption}
\usepackage{subcaption}
\usepackage{algorithm}
\usepackage{algpseudocode}

\usepackage[utf8]{inputenc} 

\title{\LARGE \bf
Rapid Online Learning of Hip Exoskeleton Assistance Preferences
}

\author{Giulia Ramella$^{1,2}$, Auke Ijspeert$^2$, and Mohamed Bouri$^{1,3}$
\thanks{Giulia Ramella received funding from the European Union's Horizon 2020 research and innovation program under the Marie-Sklodowska-Curie Grant Agreement No. 945363.}
\thanks{$^1$ Giulia Ramella and Mohamed Bouri are with the REHAssist group, Ecole Polytechnique Federale de Lausanne (EPFL). 
        {\tt\small giulia.ramella@epfl.ch}}%
\thanks{$^2$ Giulia Ramella and Auke Ijspeert are with the BioRobotics Laboratory, Ecole Polytechnique Federale de Lausanne (EPFL).}%
\thanks{$^3$ Mohamed Bouri is with the TNE Laboratory, Institute of Neuro-X, Ecole Polytechnique Federale de Lausanne (EPFL).}%
}

\begin{document}
\maketitle
\thispagestyle{empty}
\pagestyle{empty}

\begin{abstract}
Hip exoskeletons are increasing in popularity due to their effectiveness across various scenarios and their ability to adapt to different users. However, personalizing the assistance often requires lengthy tuning procedures and computationally intensive algorithms, and most existing methods do not incorporate user feedback. In this work, we propose a novel approach for rapidly learning users' preferences for hip exoskeleton assistance. We perform pairwise comparisons of distinct randomly generated assistive profiles, and collect participants preferences through active querying. Users' feedback is integrated into a preference-learning algorithm that updates its belief, learns a user-dependent reward function, and changes the assistive torque profiles accordingly. 
Results from eight healthy subjects display distinct preferred torque profiles, and users' choices remain consistent when compared to a perturbed profile. A comprehensive evaluation of users' preferences reveals a close relationship with individual walking strategies. The tested torque profiles do not disrupt kinematic joint synergies, and participants favor assistive torques that are synchronized with their movements, resulting in lower negative power from the device. This straightforward approach enables the rapid learning of users preferences and rewards, grounding future studies on reward-based human-exoskeleton interaction.
\end{abstract}

\vspace{-1em}
\section{Introduction}
Exoskeletons for walking assistance have experienced a remarkable growth in recent years. Initially developed for rehabilitation applications, these technologies are now used to partially assist healthy individuals across a wide range of contexts. As devices have evolved, various control strategies have been developed to enhance their functionality and performance. These control laws are designed to complement users' movements while minimizing constraints and preserving natural motions. In this context, tailoring the exoskeleton assistance to each user is essential to accommodate each individual’s unique movement patterns and needs. To personalize the exoskeleton support, existing contributions have successfully implemented human-in-the-loop optimization procedures with metabolic cost estimators \cite{kim_human---loop_2017, zhang_human---loop_2017, kim_bayesian_2019, franks_comparing_2021, bryan_optimized_2021, kim_reducing_2022, gordon_human---loop_2022}. However, these techniques require long tuning times and computationally expensive algorithms. On the other hand, recent studies have begun exploring novel algorithms that incorporate individual preferences to design assistive profiles driven by users' choices \cite{tucker_human_2020, tucker_preference-based_2020, ingraham_leveraging_2023, lee_user_2023}.

\begin{figure}[t]
\includegraphics[width=\linewidth]{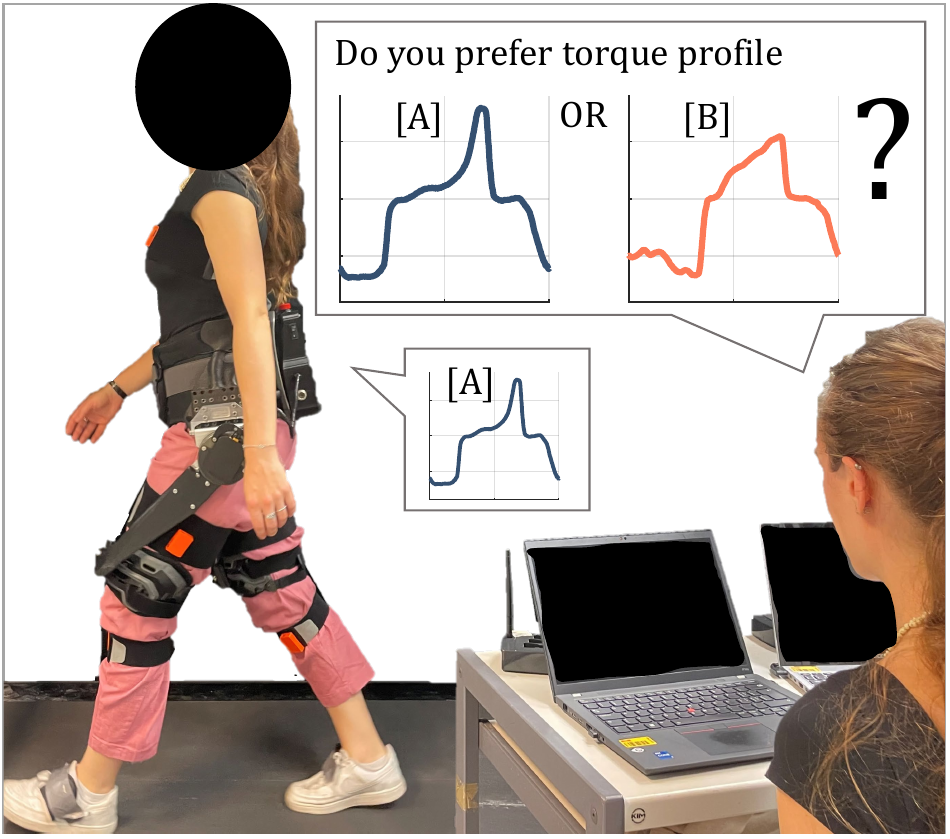}
\caption{Framework for the rapid online learning of preferred assistive torque provided by eWalk, a hip exoskeleton for partial assistance. Participants are actively queried on their preferred assistance through pairwise comparisons of distinct torque profiles. Their feedback is integrated into a preference-based algorithm which updates the belief distribution, learns a user-specif reward, and adjusts the torque profile in real-time.}
\label{FirstPage}
\centering
\vspace{-2.25em}
\end{figure}

Understanding users' preferred motion is fundamental to achieve a smooth and synergistic human-robot interaction. The field of collaborative robotics presents various learning techniques designed to capture users' preferred settings and learn the corresponding reward function. Some studies have used learning from demonstration approaches, where data is passively collected from human demonstrators to understand and predict users' preferences in robot-assisted scenarios \cite{ang_learning_2018, ravichandar_robot_2019, zhao_learning_2023, van_der_spaa_simultaneously_2024}. These methods are highly effective, but they are computationally expensive, and the learned model may overfit to specific demonstrations. Alternatively, other investigations prefer to use preference-learning algorithms for their ability to directly incorporate user feedback into the learning process, while relying on minimal data. These methods capture users preference by updating a Bayesian likelihood function after pairwise comparisons of distinct samples \cite{chu_preference_2005}. These algorithms are used for various human-robot interactions, such as manipulator tasks \cite{maccarini_preference-based_2022, zhao_learning_2023}, social navigation \cite{keselman_optimizing_2023-1}, or robot cooperation \cite{wilde_improving_2020, chappuis_learning_2024-1, prasad_learning_2021}. 

Preference-learning approaches have also been applied to lower-limb wearable devices, and recent contributions have emphasised the importance of incorporating user preferences into the development of control strategies \cite{tucker_human_2020, tucker_preference-based_2020, li_roial_2021, thatte_sample-efficient_2017, caputo_robotic_2022, ingraham_role_2022, shepherd_comparing_2020, clites_understanding_2021}. These approaches allow to capture the multifaceted nature of exoskeleton comfort in a quantifiable manner \cite{ingraham_role_2022}. Previous studies with the Atalante exoskeleton have implemented preference-based algorithms to adjust the assistive torque profiles for different users. Techniques like dueling bandits and co-active learning were employed to capture users' preferences and personalize the assistance \cite{tucker_human_2020, tucker_preference-based_2020}. In another work, an evolutionary algorithm was used to learn users' preferred settings when walking with an active ankle exoskeleton \cite{lee_user_2023}. In this approach, a pre-trained model was optimized for each participant through pairwise comparisons. Another study approached the problem from a different perspective, exploring the correlation between available biomechanical data and user-preferred ankle prosthesis stiffness. The authors used various machine learning techniques, and compared their accuracy in estimating users' preferences \cite{shetty_data_2022}. Another crucial point that has emerged in the current literature is the importance of developing algorithms relying on minimal datasets, especially given the limited data available during online optimizations \cite{le_transfer_2024, orhan_real-time_2024, grazi_kinematics-based_2022}. 

In summary, preference-learning algorithms offer the possibility to translate user preferences into quantifiable metrics and adjust wearable devices assistance accordingly \cite{ingraham_leveraging_2023}. However, current approaches require extensive pre-experiment simulation training and rely on pre-trained human models. Moreover, the current exoskeleton literature is missing preference-based methods capable of rapidly updating belief distributions in real time and learning personalized reward functions. To tackle this challenge, we propose a method for rapidly learning user preferences for the assistance provided by the hip exoskeleton eWalk. We perform pairwise comparisons of distinct torque profiles, and we actively query users on their preferred option (Figure \ref{FirstPage}). The collected feedback is incorporated into a preference-based algorithm that updates the belief distribution and adapts the assistive profile accordingly. During the comparisons, we hypothesize that users select assistive torques aligned with their walking patterns and synchronized with their movements, resulting in lower negative power from the device. Upon completion of the comparisons, the algorithm learns an individualized reward function without the need of pre-trained models, offline simulations, or extensive datasets. 

This paper is organized as follows. First, we present the learning framework and its implementation. Next, we present the experimental evaluation conducted with eight healthy subjects, and outline the metrics applied to the collected data. Then, we provide a comprehensive analysis of users preferences and the corresponding walking strategies.



\section{Methods}
In this section, we describe the torque profile provided the hip exoskeleton eWalk, which is parameterized with gait-cycle (GC) dependent features. Next, we outline the algorithm used to learn preferred torque profiles and reward functions. Afterwards, we detail the experimental protocol designed to test this approach with eight healthy subjects. 

\subsection{Torque Profile Parametrization}
For the assistance provided by the eWalk hip exoskeleton, we designed a parameterized, gait-phase-dependent torque profile with features that can vary within predefined intervals. This design ensures flexible adjustment of the torque profile shape, making it suitable for real-time adaptation. The chosen torque profile features and their range of values are presented in Table \ref{tab1} and displayed in Figure \ref{Methods}A.

\begin{table}[tbp]
\vspace{+1em}
\caption{Torque profile features}
\begin{center}
\vspace{-1em}
\begin{tabular}{c c c c}
\textbf{\textit{Feature name}}& \textbf{\textit{Range}}& \textbf{\textit{Units}}&
\textbf{\textit{Symbol}}\\
\hline
extension peak torque & [5.0, 8.0] & $Nm$ & $f_{1}$\\
extension peak time & [10.0, 20.0] & $\%GC$ & $f_{2}$\\
extension rise time & [10.0, 20.0] & $\%GC$ & $f_{3}$\\
flexion peak torque & [5.0, 8.0] & $Nm$ & $f_{4}$\\
flexion peak time & [55.0, 65.0] & $\%GC$ & $f_{5}$\\
flexion rise time & [10.0, 20.0] & $\%GC$ & $f_{6}$ \\
\hline
\end{tabular}
\label{tab1}
\end{center}
\vspace{-3.25em}
\end{table}

In this study, we selected peak torque values that ranged between 5 and 8 $Nm$ in order to guarantee comfort across all the tested torque profiles. The intent was to avoid that variations of torque profiles, if applied with high amplitudes, could cause discomfort, or induce a marching gait \cite{kang_effect_2019}. 

For each set of torque profile features, the exoskeleton controller interpolates points across the gait cycle using smooth trapezoidal interpolation curves. At every time instant, depending on the detected phase of the gait cycle, the controller provides assistance according to the corresponding value from the interpolated curve. The beginning of each gait cycle and its phases are identified using an algorithm previously developed in our laboratory \cite{manzoori_gait_2023}. 

During the experiment, various torque profiles were presented to users to gather feedback on different assistive torques and collect their preferences. To generate different torque profile shapes, the features were discretized with decimal precision and randomly sampled from their respective ranges of possible values. This process resulted in a batch of 40 randomly generated parameterized assistive torques. The number of generated torque profiles and the discretization of features were determined based on findings from a preliminary pilot study. These values were chosen to ensure that the assistive torques tested during comparisons were distinguishable to the human user. Generating additional profile options would likely not have improved the algorithm, as human subjects cannot distinguish between profiles with highly similar features. An example of torques tested with a subject is shown in Figure \ref{Methods}C.

\vspace{-0.5em}
\subsection{Learning Users' Preference}
\begin{figure}[tbp]
\includegraphics[width=\linewidth]{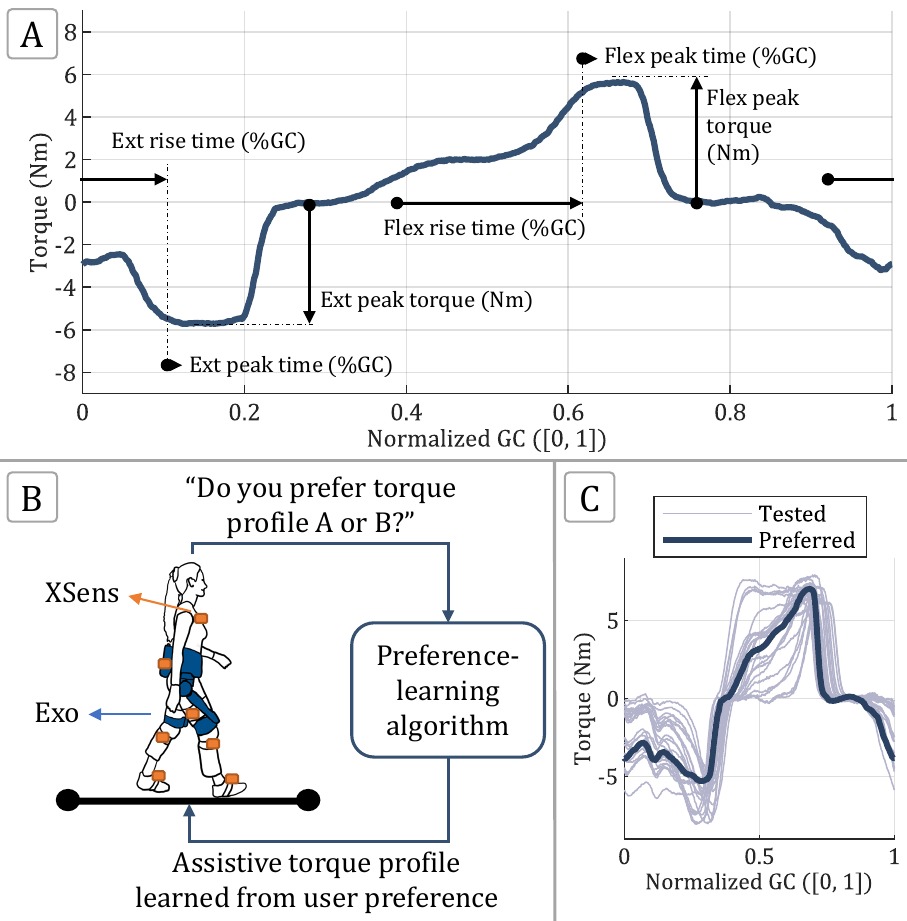}
\caption{(A): Hip torque profile parameterized using six features, which were adjusted in real-time by the preference-learning algorithm.
(B): Schematic diagram of our experiment, where users wear the eWalk hip exoskeleton and the motion capture system XSens. Active querying is used to collect user's preference between consecutive pairs of torque profiles. Individual preferences are integrated into our preference-learning algorithm, that updates the human's belief distribution and adapts the torque profile.
(C): Set of torque profiles tested on a representative subject, and the final chosen preferred assistive torque (bold blue line).}
\label{Methods}
\centering
\vspace{-2.25em}
\end{figure}

Different assistive profiles were iteratively proposed to participants in order to capture their preferred assistance. We used pairwise comparisons with active querying, where the user is actively choosing the preferred option over two distinct assistive torque profiles (``Do you prefer A or B?''). The algorithm is initialized with the set of randomly generated torque profiles, a dummy query, and an initial belief distribution. At each iteration, the user tests two distinct profiles one after the other for the same amount of time, and selects the preferred option. The response is used to update the belief distribution $p(\theta)$, adjust the weights of the reward function, and optimize the subsequent query. 
At the end of the comparisons, the algorithm outputs the final belief distribution and assistive torque profile learned from the user's preference (Algorithm \ref{alg:preference_learning} and Figure \ref{Methods}B).

To integrate human preference into the learning procedure, we developed a dedicated framework within a custom Gym environment. At each iteration, the algorithm uses a preference-based learning approach to derive a reward function from the user's feedback. The reward function ($R(\xi)$) used in this study is designed to be a linear combination of the torque profile features, and it can be expressed in the form: 
$R(\xi) = w^\top \Phi(\xi),$
where $w = [w_{1} \ w_{2} \ w_{3} \ w_{4} \ w_{5} \ w_{6}]$ 
are the weights and $\Phi(\xi)$ 
denotes the torque profile features $(f_{1},...,f_{6})$. During the learning procedure, the weights have to be learned based on the user's preference, and the features are iteratively chosen from the initial batch of torque profiles and sampled based on the user's feedback. 

As other learning algorithms, we assumed a soft-max human response model in the form 
$P(c=\xi_c) = \frac{\exp{(R(\xi_c))}}{\sum_{j}{\exp(R(\xi_j))}}$, where $P(c=\xi_c)$ is the probability that the user selects the option $\xi_c$ (``c'' as chosen) over all the ``j'' options from the set.
To update the user's belief distribution, we employed a sampling-based posterior model driven by the user's responses. The Metropolis-Hastings algorithm implemented in APReL library \cite{biyik_aprel_2022-1} is used to update the belief and the reward weights after each pairwise comparison. At each iteration, a new sample is proposed based on previous ones, and its acceptance depends on a probability ratio derived from the target distribution. The proposed sample is accepted if it improves the likelihood, otherwise it is accepted with a probability proportional to the likelihood difference. 

\begin{algorithm}
\caption{Preference Learning via Pairwise Comparisons}\label{alg:preference_learning}
\begin{algorithmic}[tbp]
    \State \textbf{Input:} Batch of torque profiles and user answer $\{A, B\}$ 
    \State \textbf{Output:} Updated belief distribution $p(\theta | \text{data})$ and user-preferred torque profile 
    \State \textbf{Initialize:} Belief distribution $p(\theta)$ and dummy query
    \While{pairwise comparison not finished}
        \State{\textbullet\ Optimize query: Present torque profile options $A$}
        \State {and $B$ to the user}
        \State{\textbullet\ User response: Ask user ``Do you prefer A or B?''}
        \State{\textbullet\ Update belief: Based on user's response, update the}
        \State{belief distribution $p(\theta | \text{data})$}
    \EndWhile
    \State \textbf{Return:} Final belief distribution and final assistive torque profile features reflecting user's preference
\end{algorithmic}
\end{algorithm}

\subsection{Experimental Protocol}
Eight healthy subjects volunteered to participate in the study (age: $27.4 \pm 3.5$ yrs old, height: $ 178.1 \pm 6.5 $ m, weight: $ 70.5 \pm 5.7 $ kg). The experimental procedures were approved by the human research ethics committee of Canton de Vaud (CER-VD) under the protocol number 2023-02305. Participants signed an informed consent prior to the evaluation, and the experiment was conducted in accordance with the principles of the Declaration of Helsinki.

The evaluation was divided into two different sessions. For both sections, subjects were instructed to walk on a treadmill at a speed of 1.1 $m/s$ while wearing the eWalk exoskeleton. In the first part, subjects familiarized themselves with the exoskeleton assistance by walking for 10 minutes with a torque profile with fixed settings. The features for this trial were kept the same for all participants, and they were $7.0 \ Nm$ for the extension and flexion peak torques, $15.0 \ \%$ GC for the extension and flexion rise times, $10.0 \ \%$ GC for the extension peak time, and $60.0 \ \%$ GC for the flexion peak time (gray line in Figure \ref{FinTorques}).  

In a second session, subjects were instructed to walk with the exoskeleton assistance, while also wearing a motion capture system. During this trial, they were asked to express their preferred assistive torque profile over 12 consecutive pairwise comparisons. For each query, they walked for 20 seconds with the first assistive torque profile, followed by 5 seconds without assistance, and then another 20 seconds with the second torque profile option. After each comparison, participants continued walking without the exoskeleton assistance while informing the experimenter of their preferred profile, before continuing with the next pairwise comparison. The timings and number of comparisons, established during a pilot study, were designed to make comparisons straightforward, avoid fatigue, and prevent users from forgetting the options due to excessively long trials. 

After a resting phase, participants underwent another round of pairwise comparisons to validate their choice of preferred assistance. This time, the preferred torque profile was compared to its perturbed version to assess if subjects still preferred their original choice. To create perturbed profiles, peak torques, and peak and rise times were changed of $ \pm 2.0 \ Nm $ and $ \pm 7.0 \ \% $ GC respectively. At each iteration, the preferred and perturbed assistive torques were proposed to users in a randomized order. 

\subsection{Measurement Systems}
The wearable device used in this study is eWalk \cite{manzoori_evaluation_2024}, an active hip exoskeleton developed by the BioRob-REHAssist group of EPFL in collaboration with the company Sonceboz. This device is designed to provide partial assistance during walking, and each side has an active degree of freedom to support hip flexion/extension, and a passive joint for hip abduction to facilitate natural walking movements. The device version used in this study weighs 5.0 $kg$ and is secured to the user's body with a pelvic corset and two strap attachments at the thighs. eWalk has two DC servo motors (Gyems RMD X8 Pro, maximum torque 32 $Nm$), and an embedded computer (BeagleBoneBlack, Texas Instruments, USA) running the control algorithm at 500 $Hz$.

The portable motion capture system XSens (XSens, Movella, The Netherlands) was used to record kinematic data. The system has a dedicated software that records and reprocesses IMU data to reconstruct the walking kinematics of the user. In this experiment, lower-limbs data was collected at a sampling frequency of 100 $Hz$, and the system was calibrated for each participant prior to data collection.

During the experiments, we used two separate computers. The first one was running the XSens software to collect kinematic data from the IMUs in real time. A second computer was connected via Wi-Fi to the exoskeleton in order to visualize data from the device in real-time (in the custom interface), and to run the learning algorithm and send the updated torque profile to the exoskeleton controller.

\subsection{Metrics}
Collected data was processed offline with custom routines in MATLAB 2023b (The MathWorks, Natick, MA, USA). Before being imported into MATLAB, XSens data was pre-processed using the dedicated software in order to extract kinematic data of the lower limbs.  We computed relevant metrics for all subjects, labeled as S1, S2, ..., S8 in the following sections.

\textbf{\textit{Algorithm data}} To investigate individual preferences, features and weights of the final preferred torque profiles were analysed for each subject. To study the distribution of preferences across users, the mean and standard deviation were computed for each feature. For all subjects, the reward function weights over the 12 pairwise comparisons were analyzed to evaluate the algorithm's learning curve. During the validation trial, we examined if users were choosing the preferred torque profile over its perturbed version.

\textbf{\textit{Kinematic data}} Joint kinematic synergies were analyzed for each participant to examine lower limb coordination. Phase plots of knee and hip flexion/extension angles were used for this purpose. Synergies were measured across the entire set of pairwise comparisons, in order to assess whether different assistive profiles altered normal walking patterns. 
Additionally, the ratio of stance time to swing time was calculated to evaluate subjects' walking strategies. The stance and swing durations were computed from the heel-strike signal extracted from the XSens software. 

\textbf{\textit{Exoskeleton data}} From the exoskeleton data, we extracted hip joint kinematics (from motor encoders) and torque profiles, and segmented the signals using the heel-strike information. The final preferred torque profile for each participant was determined as the mean torque applied to the legs. The duration of the torques was normalized to the [0, 1] interval to facilitate inter-subject comparison. 
The power profiles ($P(t)$) were derived by segmenting the exoskeleton signals into individual gait cycles, and then calculating the product of the hip joint angular velocity  ($\omega(t)$, rad/s) and the commanded torques ($\tau(t)$, Nm), using the formula 
$ P(t) = \omega(t) \cdot \tau(t), \text{for } 0 \leq t \leq T_{\text{gait cycle}} $. 

A power ratio (PR) metric was calculated for each torque profile tested by users. This metric is defined as the unsigned ratio between the mean negative power ($P_{\text{-,mean}}$) and the mean positive power ($P_{\text{+,mean}}$) during the gait cycle. Efficient power transmission from the exoskeleton is achieved when the generated torque is applied synchronous with the user's movements, maximizing the positive power output. Therefore, the PR metric provides insight into how much of the generated power is being dissipated or lost (negative values), versus how much is contributing to productive power (positive values) during the human-robot interaction. The PR can be expressed as:
$
\text{PR} = \frac{P_{\text{-,mean}}}{P_{\text{+,mean}}} = \frac{\text{mean}|(\tau \cdot \omega)_{\text{negative}}|}{\text{mean}(\tau \cdot \omega)_{\text{positive}}}
$

This metric was computed for each torque proposed to subjects during pairwise comparisons, considering the average of each tested assistive torque. This approach enabled a comparison between the average PR of the torque profiles that were selected, versus the average of those that users discarded.

\section{Results \& Discussion}

\begin{figure}[tbp]
\includegraphics[width=\linewidth]{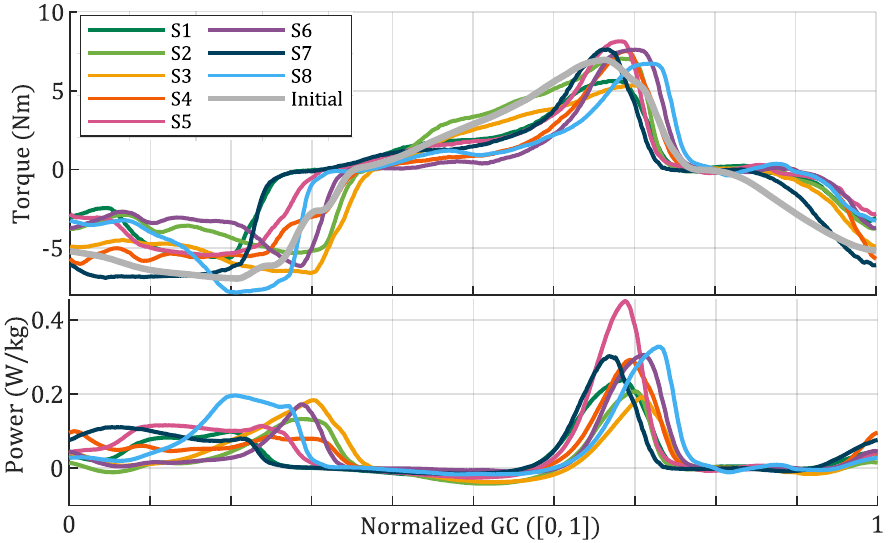}
\caption{Preferred assistive torque profiles (above), and corresponding power profiles (below), at the end of the pairwise comparisons, for each participant to the experiment. The ``Initial'' torque profile is used during the first session of familiarization with the exoskeleton, and kept the same across subjects.}
\label{FinTorques}
\centering
\vspace{-2em}
\end{figure} 

\begin{figure}[tbp]
\includegraphics[width=\linewidth]{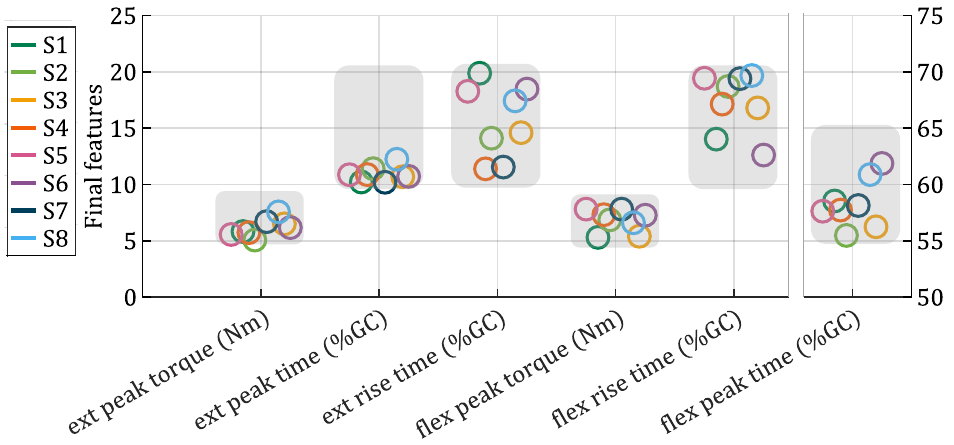}
\caption{Final preferred torque profile features at the conclusion of the learning process for each subject. The gray areas represent the possible range of values for each feature.}
\label{FinalFeat}
\centering
\vspace{-1.5em}
\end{figure}

With our preference-learning framework, we learned users preferences and reward functions for the assistive torque profiles provided by the hip exoskeleton eWalk. Users exhibited diversified preferred torque profiles, and we analyzed their individual behaviors along with the computed metrics.

\textbf{\textit{Algorithm data}} At the end of the queries, participants showed diversified features corresponding to their preferred assistance (Figure \ref{FinalFeat}). The greatest variations were observed in the extension rise time, as well as the flexion peak and rise time values. Across users, there is a clear preference for extension peak time values at the lower end of the interval (Table \ref{tab2} and Figure \ref{FinalFeat}). 

\begin{table}[tbp]
\vspace{+1em}
\caption{Distribution of final torque profile features across subjects.}
\begin{center}
\vspace{-1em}
\begin{tabular}{c c c}
\textbf{\textit{Feature name}}& \textbf{\textit{Mean}}& \textbf{\textit{Std}} \\
\hline 
extension peak torque & 6.1 & 0.8 \\
extension peak time & 10.9 & 0.7 \\
extension rise time & 15.7 & 3.3 \\
flexion peak torque & 6.8 & 1.0 \\
flexion peak time & 17.2 & 2.7 \\
flexion rise time & 58.3 & 2.2 \\
\hline
\end{tabular}
\label{tab2}
\end{center}
\vspace{-1.8em}
\end{table}

\begin{figure}[tbp]
\includegraphics[width=\linewidth]{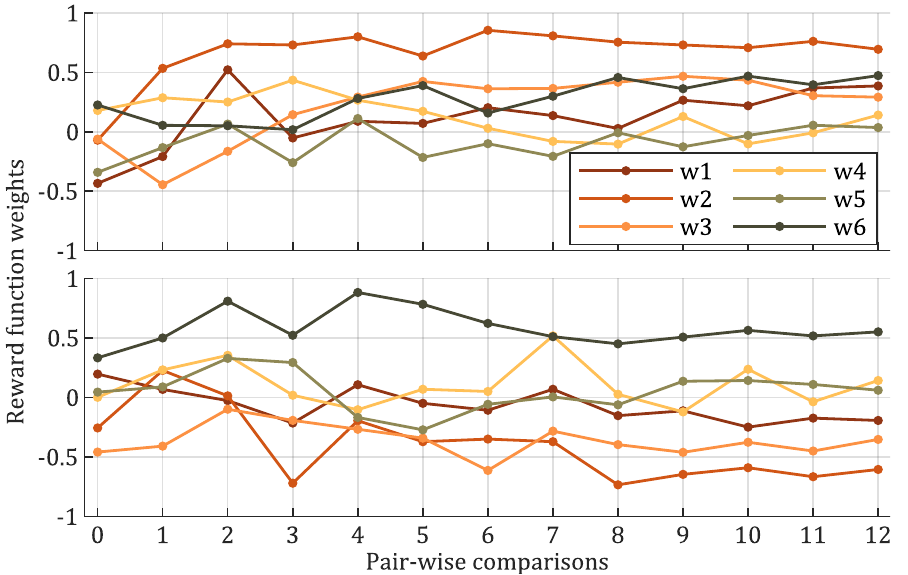}
\caption{Normalized weights of the reward function over the pairwise comparisons for two representative subjects (S2 above and S3 below). At the beginning, the weights are initialized with random values. The algorithm progressively learns the values based on user's feedback.}
\label{Weights}
\centering
\vspace{-1em}
\end{figure}

\begin{figure}[tbp]
\includegraphics[width=\linewidth]{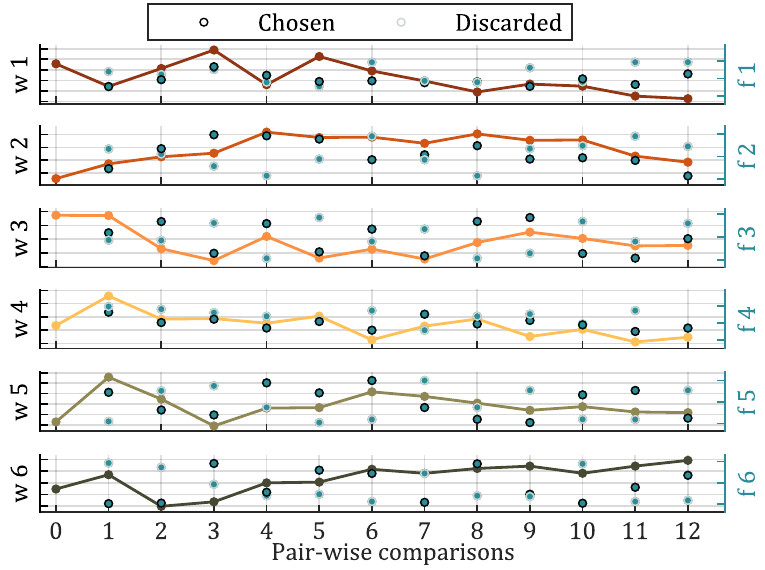}
\caption{Pairs of torque features proposed to participants at each comparison (right axis). Users' choices updated the weights of the reward function (left axis) over the pairwise comparisons (representative subject, S4).}
\label{Weightsfeats}
\centering
\vspace{-1.75em}
\end{figure}

At the beginning of the learning process, the weights of the reward function were variable, reflecting the algorithm's exploration of different possibilities (representative subjects in Figure \ref{Weights}). As the algorithm processed, the weights gradually stabilized around their final values. As illustrated in Figure \ref{Weightsfeats}, the learning procedure  updated the reward weights at each iteration based on the user's selection between the two proposed torque profile features.

During the validation phase, most subjects consistently selected their preferred torque profile over the perturbed version. However, for four participants, validation was unsuccessful when perturbing the rise times. This result suggests that when peak time and peak torque are synchronized with individual preferences, users are less sensitive to variations in the timing required to reach those peaks. The importance of peak timings in personalized assistance was also underlined in a recent study \cite{bryan_pilot_2024}.

\begin{figure}[tbp]
\includegraphics[width=\linewidth]{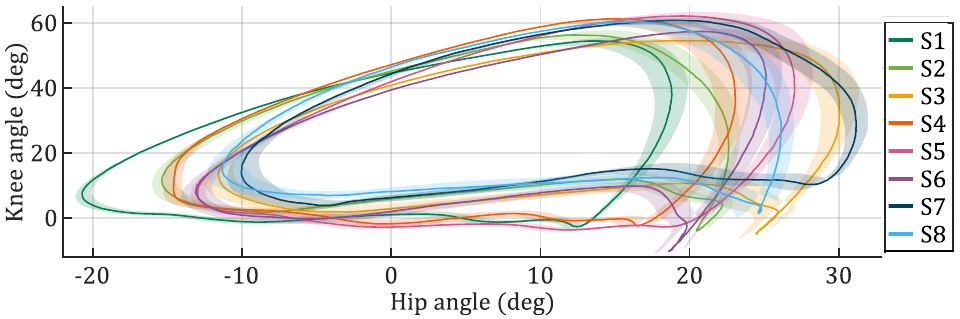}
\caption{Kinematic joint synergies for each subject. The phase plots illustrates the hip (deg) and the knee (deg) flexion-extension angles during the comparisons of different torque profiles.}
\label{KinSynergies}
\centering
\vspace{-1.25em}
\end{figure}

\textbf{\textit{Kinematic data}} When comparing subjects, kinematic joint synergies exhibited distinct patterns depending on the unique walking strategy adopted by each participant (Figure \ref{KinSynergies}). When focusing on each user, individual kinematic synergies displayed minimal variations, 
implying that different assistive torque profiles were not altering users' gait pattern. Preserving individual walking strategies has long been recognized as a crucial factor to ensure a synergistic human-exoskeleton interaction \cite{lewis_invariant_2011}. 

\begin{figure}[tbp]
\vspace{+1em}
\includegraphics[width=\linewidth]{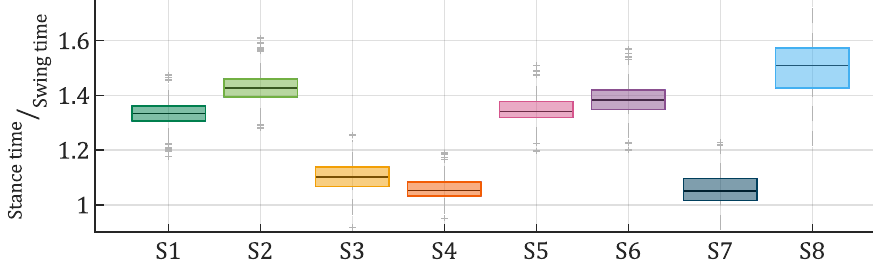}
\caption{The stance-to-swing time ratio for each subject during the pairwise comparisons of torque profiles.}
\label{StanceSwing}
\centering
\vspace{-2em}
\end{figure}

The ratio of stance and swing times varied among users, reflecting their personal gait patterns. For each subject, the ratio stayed consistent throughout the comparison trials, with minor variations possibly 
experienced when the exoskeleton assistance was activated. Two subgroups emerged from the data (Figure \ref{StanceSwing}): S3, S4, and S7 maintained a ratio close to 1.0, indicating balanced durations between the stance and swing phases. The other subjects exhibited higher ratios, indicating a longer time spent in the stance phase. While longer stance durations are linked to increased stability \cite{winter_human_1995}, the duration of walking phases changes in case of variations from the preferred walking speed \cite{fukuchi_effects_2019}. In our experiment, since we used a constant walking velocity on the treadmill, users adjusted their gait patterns accordingly.

\textbf{\textit{Exoskeleton data}} At the end of the comparisons, subjects' choices resulted in diversified preferred torque profiles and corresponding power profiles (Figure \ref{FinTorques}). For all users, the final assistive torque differed from the initial torque used during the familiarization phase. This result indicates that, despite being familiarized with a predefined assistance, subjects preferred to actively adjust the torque profile in real-time. These results are consistent with previous studies with ankle exoskeletons, where the preferred assistive torque profile varied among participants based on their individual preferences \cite{ingraham_leveraging_2023, lee_user_2023}. Additionally, all users chose extension peak times at the lower end of the interval, resulting in profiles where the extension assistance began before the heel strike. As displayed in Figure \ref{FinTorques}, preferred assistive torques started at around 90 $\%$ of the previous gait cycle. This behavior is consistent with findings from other studies, where early initiation of optimized assistive torques was beneficial to users \cite{gordon_human---loop_2022, franks_comparing_2021, bryan_optimized_2021}.

Further analysis confirmed our initial hypothesis that individual walking strategies influenced the selection of preferred torque profiles. To reveal this trend, we performed a subject-wise comparison of the chosen assistive profile (Figure \ref{FinTorques}) and the corresponding stance-to-swing ratio (Figure \ref{StanceSwing}). S3, S4, and S7 (with smaller stance/swing ratio) selected profiles with a lower extension rise time, leading to assistive torques that provided sustained support during hip extension. On the other hand, S8 (with largest stance-to-swing ratio) chose a preferred assistance with large peak and rise times during extension, resulting in a shorter but more pronounced stance assistive torque. While swing assistive profiles appear similar across participants, a detailed analysis reveals different choices in timing and peak values.

\begin{figure}[tbp]
\includegraphics[width=\linewidth]{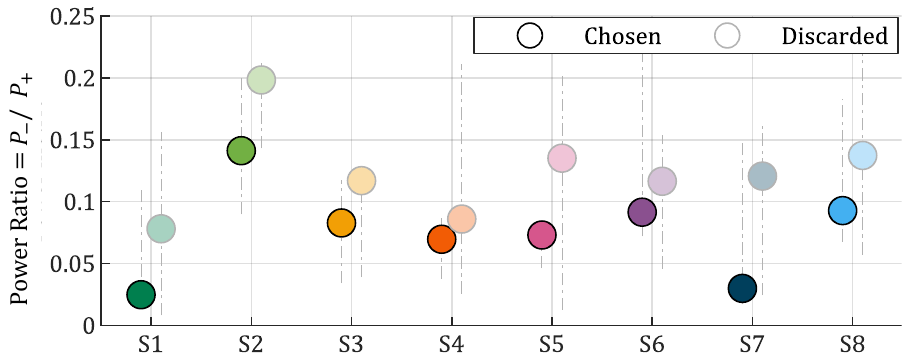}
\caption{Mean Power Ratio (PR) for each subject during pairwise comparisons, computed as the unsigned ratio of the mean negative power over the mean positive power. For each user, the left filled dot represents the PR of chosen profiles, while the right shaded dot indicates the PR of the discarded assistive profiles.}
\label{FinalPower}
\centering
\vspace{-2.25em}
\end{figure}

The power transmitted from the exoskeleton ($P=\omega \cdot \tau$, as defined in the Methods) represents an indicator of how well the power is delivered: positive power is linked to efficient energy transmission, while negative values indicate power loss during the human-exoskeleton interaction \cite{lim_parametric_2023}. The literature widely agrees on the importance of ensuring efficient power transmission from the exoskeleton to the user \cite{quinlivan_assistance_2017, bryan_optimized_2021, kim_reducing_2022}. 
Consistently with existing works, we found that users chose torque profiles with smaller losses in the power transmission from the exoskeleton. For each user, Figure \ref{FinalPower} illustrates the mean of the PR metric for chosen torque profiles (on the left, darker color) compared to the mean PR of the discarded torques (on the right, lighter shade). As we hypothesized, users consistently selected torque profiles that minimized negative power in relation to positive power. Conversely, they generally rejected assistive profiles with higher negative-to-positive power ratio. 
Furthermore, it is well-established that both the timing and amplitude of assistance play a crucial role in the positive power generated by exoskeletons \cite{lee_effects_2017, young_biomechanical_2017, bryan_optimized_2021}. This aligns with our observations, as variations in these parameters across the different torque profiles had a significant impact on the power output. Although maximizing positive power was not the primary focus of our learning algorithm, the results suggest that it may be considered as an important factor in future research.

\vspace{-1em}
\section{Conclusion}

In this work, we present a novel method to rapidly learn user preferences for the assistive torque of the hip exoskeleton eWalk. We generated random batches of parametrized torque profiles, with features varying within predefined intervals. These profiles were tested on healthy users walking on a treadmill. Through consecutive pairwise comparisons, we actively queried participants on their preferred assistive profile, and integrated this feedback into a preference-learning algorithm. After each comparison, the algorithm updated the user’s belief distribution, adapted the torque profile features, and learned a user-specific reward function. Our method rapidly derived rewards from user feedback, without the need for large datasets, pre-trained models, or re-training of simulations. These findings will ground future studies centered on reward-based human-exoskeleton interaction.

We learned torque profile preferences for eight healthy exoskeleton users, and their choices remained consistent when compared to perturbed variations of the preferred profiles. We performed a comprehensive analysis to examine relationships among users preferences, walking kinematics, and efficiency in power transmission. Results show tied relations between the stance phase duration and the preferred assistive torques. While the tested torque profiles did not alter joint kinematics, participants exhibited a distinct preference for torque profiles that limited negative power transmissions from the exoskeleton. Our analysis presents some limitations, as it was conducted at a single walking speed, walking energetics were not measured, and the reward function was assumed to be linear in the features. These factors will be addressed in our future research. 

In conclusion, we present a straightforward online approach to learn user preferences and rewards for the assistive torque profiles provided by the hip exoskeleton eWalk. This method enables rapid personalization of exoskeleton assistance, advancing the development of customized control strategies.

\newpage

\bibliographystyle{ieeetr} 
\bibliography{PL} 

\end{document}